\documentclass[sigconf]{acmart}
\usepackage{wrapfig}
\newcommand{\tool}[0]{\textsc{EnergyVis}}
\AtBeginDocument{%
  \providecommand\BibTeX{{%
    \normalfont B\kern-0.5em{\scshape i\kern-0.25em b}\kern-0.8em\TeX}}}

\copyrightyear{2021}
\acmYear{2021}
\setcopyright{acmlicensed}\acmConference[CHI '21 Extended Abstracts]{CHI Conference on Human Factors in Computing Systems Extended Abstracts}{May 8--13, 2021}{Yokohama, Japan}
\acmBooktitle{CHI Conference on Human Factors in Computing Systems Extended Abstracts (CHI '21 Extended Abstracts), May 8--13, 2021, Yokohama, Japan}
\acmPrice{15.00}
\acmDOI{10.1145/3411763.3451780}
\acmISBN{978-1-4503-8095-9/21/05}

\acmSubmissionID{1283}

\begin{document}

\title{\tool{}: Interactively Tracking and Exploring \\ Energy Consumption for ML Models}

\author{Omar Shaikh}
\affiliation{%
  \institution{Georgia Institute of Technology}}
\email{oshaikh@gatech.edu}

\author{Jon Saad-Falcon}
\affiliation{%
  \institution{Georgia Institute of Technology}}
\email{jonsaadfalcon@gatech.edu}

\author{Austin P Wright}
\affiliation{%
  \institution{Georgia Institute of Technology}}
\email{austinpwright@gatech.edu}

\author{Nilaksh Das}
\affiliation{%
  \institution{Georgia Institute of Technology}}
\email{nilakshdas@gatech.edu}

\author{Scott Freitas}
\affiliation{%
  \institution{Georgia Institute of Technology}}
\email{safreita@gatech.edu}

\author{Omar Isaac Asensio}
\affiliation{%
  \institution{Georgia Institute of Technology}}
\email{asensio@pubpolicy.gatech.edu}

\author{Duen Horng (Polo) Chau}
\affiliation{%
  \institution{Georgia Institute of Technology}}
\email{polo@gatech.edu}

\renewcommand{\shortauthors}{Shaikh et al.}

\begin{abstract}
    The advent of larger machine learning (ML) models have improved state-of-the-art (SOTA) performance in various modeling tasks, ranging from computer vision to natural language. As ML models continue increasing in size, so does their respective energy consumption and computational requirements.
    However, the methods for tracking, reporting, and comparing energy consumption remain limited. 
    We present \tool{}, an interactive energy consumption tracker for ML models. 
    Consisting of multiple coordinated views, 
    \tool{} enables researchers to interactively track, visualize and compare model energy consumption across key energy consumption and carbon footprint metrics (kWh and CO$_2$), helping users explore alternative deployment locations and hardware that may reduce carbon footprints. \tool{} aims to raise awareness concerning computational sustainability by interactively highlighting excessive energy usage during model training; and by providing alternative training options to reduce energy usage. 
\end{abstract}




\begin{teaserfigure}
\begin{center}
  \includegraphics[width=.8822\textwidth]{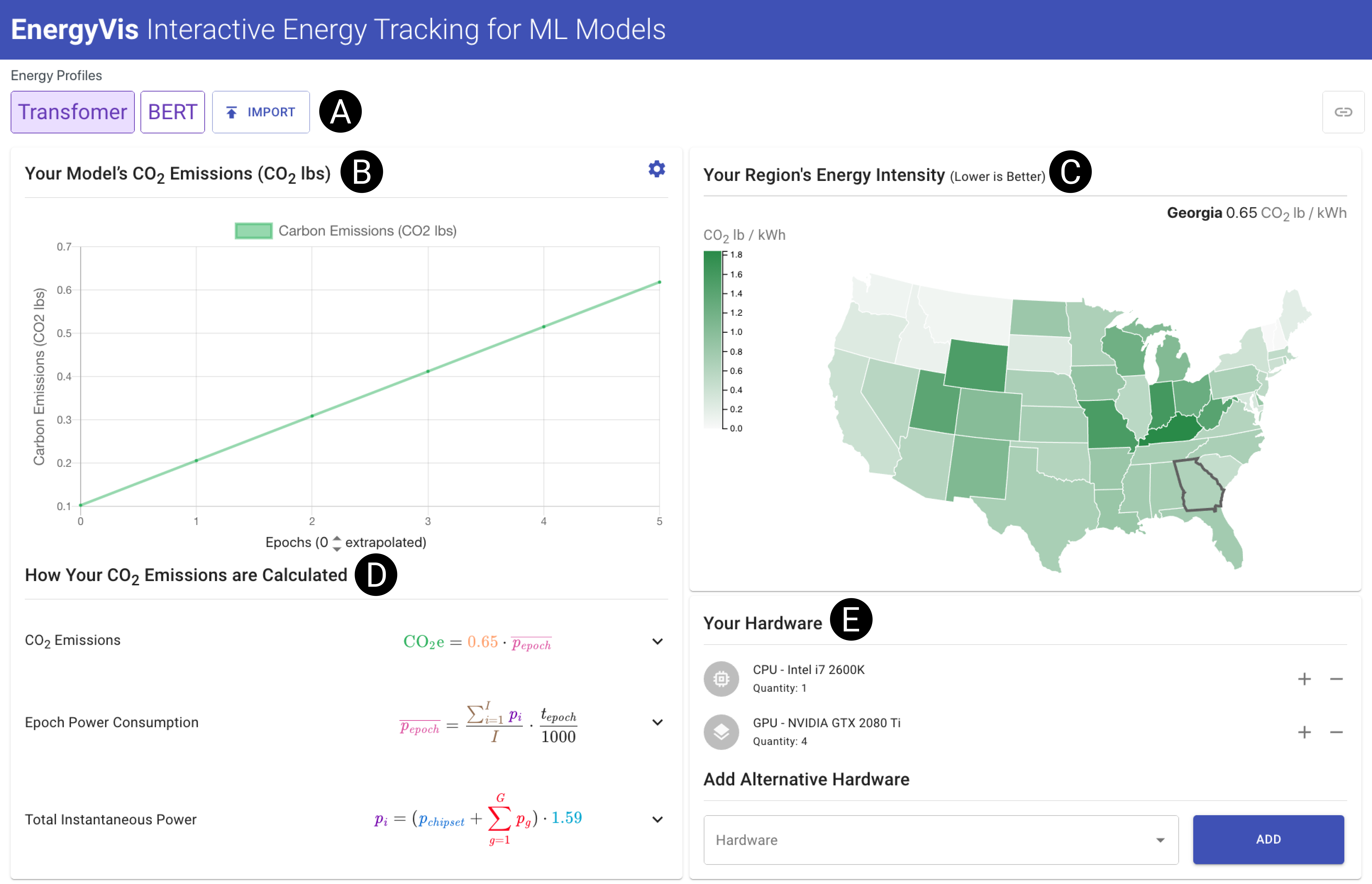}
  \caption{The \tool{} user interface, with multiple coordinated views. 
  \textbf{(A)} The \textit{Model Energy Profile View} allows users to select an energy profile of pre-loaded models, generate new profiles (for models that a user wishes to train), and import saved profiles. \textbf{(B)} The \textit{Consumption Chart} allows users to view the energy and carbon consumption of their selected model. \textbf{(C)} Using the \textit{Model Region} view, users can view the region where a model was trained, and select regions with a lower energy intensity as an alternative to reduce emissions. \textbf{(D)} Users can expand the \textit{Colored Equation}s for succinct descriptions of various variables and how they contribute to calculating a model's emissions. \textbf{(E)} Finally, users can view or adjust hardware used to train a model using \textit{Alternative Hardware}. 
  \Description[A teaser figure highlighting the components of EnergyVis]{The teaser figure for EnergyVis, outlining the multiple coordinate views that make up the interface.}
} 
  \label{fig:teaser}
\end{center}
\end{teaserfigure}
\maketitle

\begin{table*}
\centering
\small %
\begin{tabular}{@{}lr@{}}  %
\toprule
Consumption Mode &  CO$_{2}$e (Tons) \\ 
\midrule
Air travel, 1 passenger, NY to SF & 0.99 \\
Human life, avg, 1 year & 5.52 \\
American life, avg, 1 year & 18.08 \\
Car, avg incl. fuel, 1 lifetime & 63.00 \\
\bottomrule
\end{tabular}
\hspace{1em}
\begin{tabular}{@{}lr@{}}  %
\toprule
Consumption Mode &  CO$_{2}$e (Tons) \\ 
\midrule
BERT$_{base}$ &  0.72\\
NLP Pipeline (parsing, SRL) & 39.23 \\
Neural Architecture Search & 313.08 \\
\bottomrule
\end{tabular}
\caption{CO$_2$ consumption from familiar sources compared to NLP pipelines, from \citet{strubell-etal-2019-energy}
}
\label{tab:consumption_normal}

\end{table*}

\section{Introduction \& Background}
Researchers and practitioners alike utilize machine learning (ML) to successfully model and draw insights from datasets. 
Continued advancements in ML have also significantly furthered modeling performance. 
However, as datasets grow more complex, researchers are switching to larger architectures that require increasingly more compute~\cite{amodei_hernandez_2018, ILSVRC15, wang2019superglue, wang-etal-2018-glue}. 
These models' parameter counts have increased by over $300,000\times$ from 2012 to 2019 \cite{amodei_hernandez_2018}. Models like GPT-3 are reflective of this trend, with performance increases relying on scaling previously successful architectures \cite{brown2020language}. 
Due to this rapid increase, ML researchers will soon be forced to account for computational efficiency for improved performance \cite{hernandez_brown_2020}. 
Other work, like \citet{schwartz2019green}, raises concerns regarding computational equity: as models grow larger and larger, computational resources make replication prohibitive. 
Pushing for Green AI, where compute requirements themselves are reduced, will guarantee a reduction in carbon footprints \cite{schwartz2019green}.

Continuing to reduce compute resources and energy consumption has several benefits. Firstly, smaller research teams and companies simply might not have access to renewable resources.
In this case, tracking just efficiency plays a key role in computational equity,
as it enables users to explore how carbon footprints change if their models were deployed elsewhere. 
Secondly, carbon credits alone are an indirect---and occasionally ineffective---approach to reducing carbon output, as they require third-party verification, offset direct responsibility, and require monetary compensation, which might be impractical for smaller labs or companies \cite{lovell2010understanding}. Thirdly, energy from training models can be repurposed towards powering homes or essential infrastructure. 
Preliminary estimates from \citet{strubell-etal-2019-energy}, summarized in Table \ref{tab:consumption_normal}, highlight the significant carbon footprints coming from various natural language processing pipelines. For example, training a state-of-the-art NLP model (BERT \cite{devlin2019bert}) on a GPU produces comparable carbon output to a trans-atlantic flight \cite{strubell-etal-2019-energy}. By directly reducing computationally intensive tasks, identified through monitoring ML training process, the computing community can help play a role in reducing emissions.

Despite these increases in computational intensity, tracking and analyzing energy usage from training/evaluating these models remains a challenge. Prior work, like \citet{henderson2020systematic}, covers the lack of systemic reporting for carbon output stemming from ML models, while providing a framework and a tool for tracking and reporting energy usage using static graphs. \citet{anthony2020carbontracker} builds on prior work by offering predictions for total carbon consumption through a command line interface. Finally, \citet{lottick2019energy} generates static energy usage reports at the end of an experiment, and includes a set of fixed, location based counterfactuals. However, these tools are limited to command line interfaces (CLI), or use static visuals to communicate results. Furthermore, practitioners may be unaware of alternatives to reduce their carbon footprint even after tracking and monitoring their models using prior work.

To help address a small subset of these challenges, we are developing \tool{}, an interactive system that tracks energy consumption for machine learning (ML) models, while providing practitioners and researchers with alternative options to reduce this consumption. By logging electricity usage from hardware components (kilowatt hours, kWh, or watts, W), \tool{} collects energy usage from various computationally intensive experiments. Using energy resource data from the National Renewable Energy Laboratory (NREL) \cite{cambium}, along with collected energy usage, \tool{} can estimate carbon output from experiments run in various locations, while exploring location based alternatives. \tool{} also aims to raise awareness with respect to computational sustainability. In this work, our contributions include:

\begin{itemize}
    \item \textbf{\tool{}}, a web-based system to interactively allow practitioners to explore energy efficiency costs (CO$_2$ consumption, kWh, etc.) associated with training ML models. Building on prior energy tracking tools, used through the command-line interface or consisting of static graphs, \tool{} provides a novel web-based interface for monitoring energy utility across experiments. Furthermore, \tool{} can optionally sync to backend training code using a Python plugin, which updates projections and recorded energy consumption with respect to a running experiment. 
    \item \textbf{Visual comparisons across experimental setups:} \tool{} provides users with the option to visually compare energy efficiency projections from other experiments, allowing for cross-model energy efficiency comparisons. Given experimental data from other work, users can define hardware and energy source estimates, \textit{overlaying their own tracking data with efficiency metrics from these works.} If possible, users can also import reports generated by \tool{}, allowing for direct comparisons instead of estimations. 
    \item \textbf{Interactive exploration of alternatives for reducing energy consumption and carbon footprint:} \tool{} allows users to pause experiments and observe the impact of alternatives on energy consumption projections. Concretely, \tool{} allows users to select hardware and location counterfactuals while watching (real-time) projections for carbon usage change appropriately. By enabling alternative exploration, users can change elements of their training process to consume less energy, and reduce their footprint.  
\end{itemize}

\section{System Design and Implementation}
\tool{} is an interactive tool that allows users to compare energy usages across model training pipeline, and find alternatives to reduce energy consumption and CO$_{2}$ production. \tool{} consists of two different modes. The first mode is a \textbf{preloaded mode}, which allows users to load results from previously live-tracked models. The second mode is a \textbf{live-tracking mode}, which allows users to actively track results from a training model. To support both of these modes, \tool{} consists of a frontend, which handles visualization, comparison, and alternatives selection. A backend, which is launched when users train a model using \tool{}'s Python library, handles live-tracking of models.

The frontend of \tool{} was written using modern web technologies like React.js and D3, and consists of several components. The \textbf{(1) Energy Profile View} component loads \textbf{model energy profiles} into \tool{}'s UI. Energy Profiles are JSON (meta-data) files that contain information about a models energy usage over training epochs. The \textbf{(2) Alternative Model Region} component highlights energy intensity (how much carbon is produced per kilowatt-hour of electricity) across different regions, and allows users to select alternate regions to train their models. Likewise, the \textbf{(3) Alternative Hardware} component displays a user's hardware and allows alternative selection. The \textbf{(4) Consumption Chart} component graphs the consumption from a model profile, along with selected alternatives. Finally, \textbf{(5) Color Equation} components display equations used to calculate a model's emissions given selected alternatives.

\begin{figure}[h]
  \centering
  
  \includegraphics[width=\linewidth]{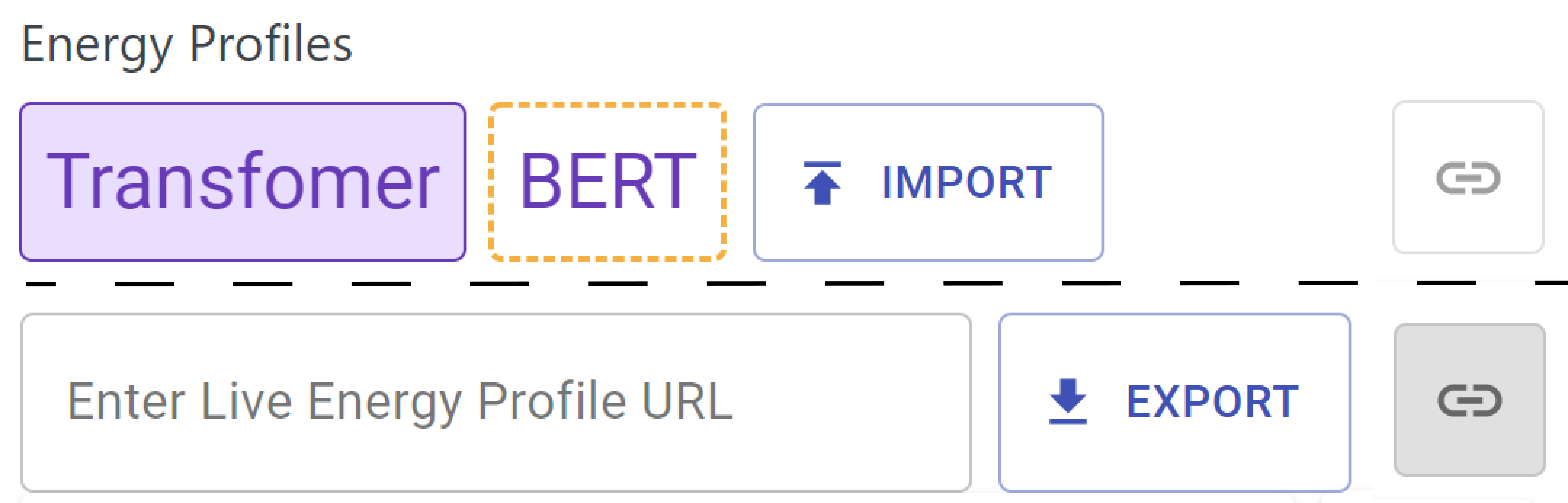}
  \caption{The Energy Profile View component allows users to switch between Energy Profiles. A preloaded mode \textbf{(top)} allows users to select, compare, or import preloaded energy profiles. Users can compare efficiency metrics across different energy profiles by right-clicking on a profile. A live tracking mode \textbf{(bottom)} allows users to collect their own energy profile for arbitrary models and export them, using \tool{}'s optional backend.}
  \Description{A figure of the EnergyProfile View, with the Transformer model selected (top), and the ability to enter a custom link for live tracking (bottom).}

  \label{fig:energyprof}
\end{figure}
\label{energy-profile-section}

\subsubsection{Model Energy Profile View}
The Model Energy Profile View allows users to select between model energy profiles. Energy Profiles are JSON metadata files containing information about a model's training pipeline. These files consist of information like the hardware used to train the model, the location the model was trained in, how much energy the model used per epoch, and how long each epoch lasted. After energy profiles are imported, they appear as a button in the profile view.

The energy profile view also allows users to switch between the \textbf{preloaded mode} and the \textbf{live tracking mode}, using the link button placed at the right of the view (seen in Figure \ref{fig:energyprof}). In the preloaded mode, users can either import an energy profile using the import button, or select from pre-imported models (in Figure \ref{fig:energyprof}, a Transformer \cite{vaswani2017attention} and BERT \cite{devlin2019bert} model are imported). Users can also right click on unselected energy profiles to load their data as alternatives in \tool{}'s UI; alternative profiles are highlighted using a dashed orange border. If users want to collect their own energy profiles, the live-tracking mode allows users to input a URL to \tool{}'s backend (described in Section \ref{backend-section}). After training a model in live-tracking mode, users can export a profile using the export button.

\subsubsection{Alternative Model Region}
\begin{figure*}[h]
  \centering
  \includegraphics[width=\linewidth]{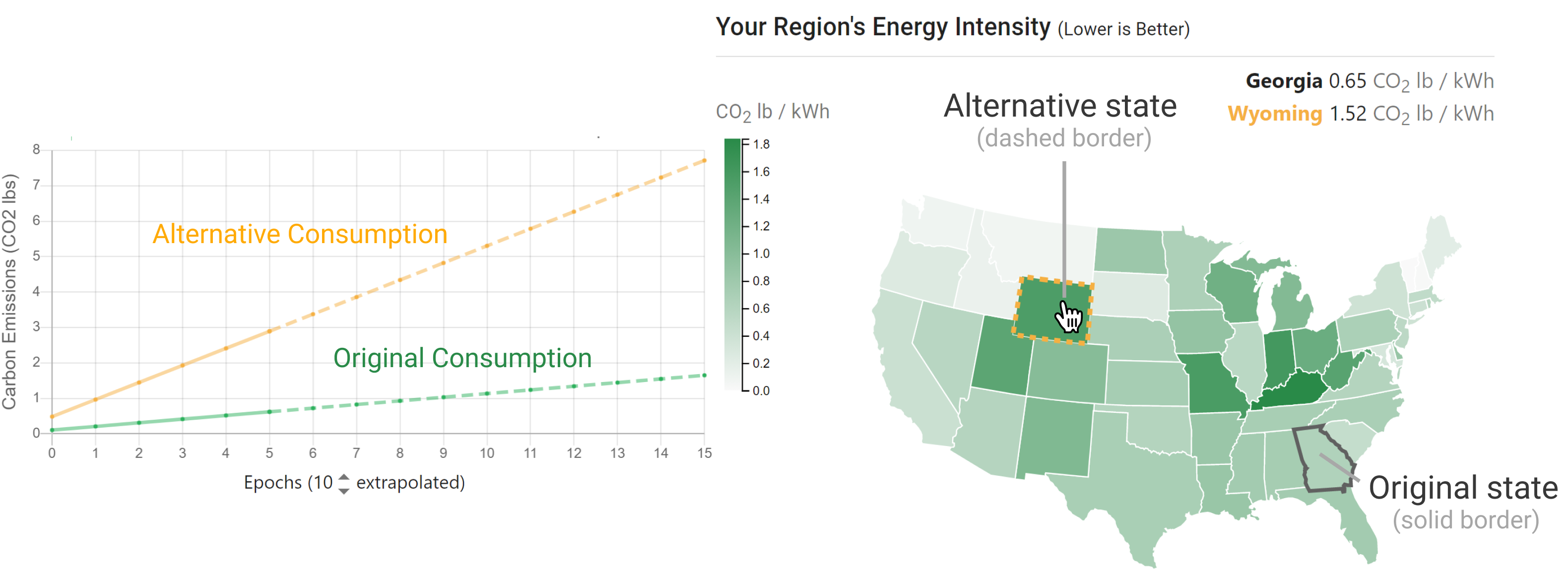}
  \caption{Adjustments made to other components (like the Energy Map, \textbf{right}) reflect as an alternative consumption line on the Consumption Chart (\textbf{left}). In this figure, a user is hovering over Wyoming, causing the alternative to be rendered alongside the original consumption line.}
  \Description{A figure highlighting how components in EnergyVis interact. A user is currently hovering over a state, and the alternative energy consumption for the hovered state is visible in the Consumption Chart component}
  \label{fig:graphmap}
\end{figure*}

The Alternative Model Region Component (Figure \ref{fig:graphmap}, right) allows users to select alternative regions for training their models. Alternative region emissions are computing using \textit{energy intensity}; some regions produce more lbs of CO$_{2}$ for the same amount of electricity (in kWh, or kilowatt hours). Energy intensity is measured in $\frac{CO_{2} lbs}{kWh}$. As different regions are selected (by clicking) or previewed (by hovering) on the map, alternative emissions are computed by substituting new energy intensities in the emission computations. This allows users to make informed decisions about deployment locations for their models.

\label{model-region}
\subsection{\tool{} Frontend}

\begin{figure}
  \begin{center}
    \includegraphics[width=0.42\textwidth]{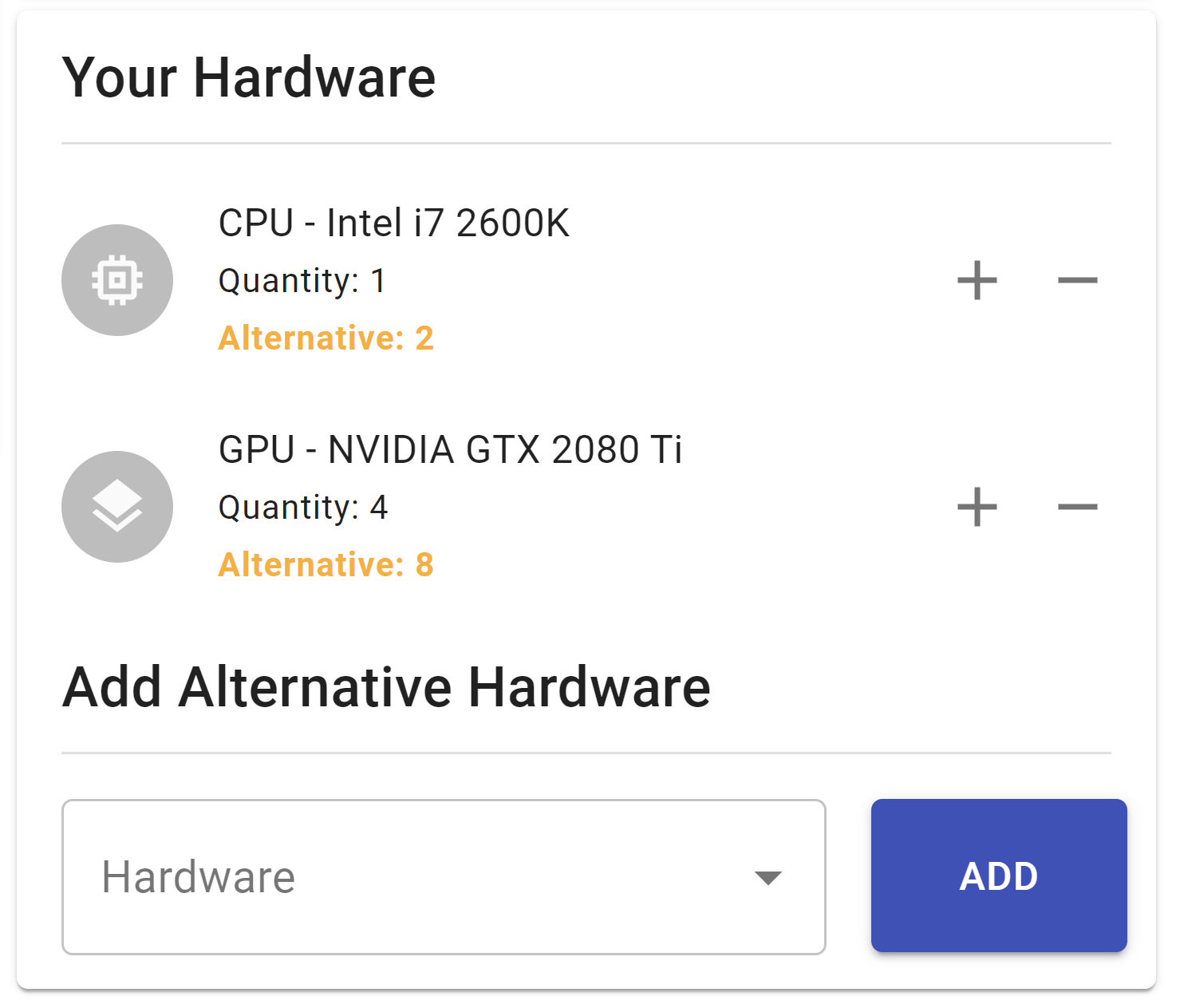}
  \end{center}
  \caption{The alternative hardware component, where users can adjust or add hardware in their current training pipeline, enabling them to view hardware based energy alternatives.}

  \Description{The alternative hardware component. In this figure, a user has selected alternative hardware quantities for their default machine. The alternatives are highlighted in a different color to distinguish them from the original components. }
  \label{fig:althardware}
\end{figure}

\subsubsection{Alternative Hardware}
\label{alternative-hardware}

\tool{} also enables users to explore hardware based alternatives. The alternative hardware component (Figure \ref{fig:althardware}) allows users to change the quantity of their hardware, or add new hardware to their training pipeline, while viewing updated energy usages. To implement hardware alternatives, we utilize a dataset from \citet{sun2020summarizing}, consisting of power draw $p$ and floating point operations per second (FLOPS) $s$ for over 4000 GPUs and CPUs. To calculate alternative power consumed over an epoch, we multiply the original power used by $\frac{p_{a}/s_{a}}{p/s}$, where $p_{a}$ and $s_{a}$ are the power and FLOPs for the selected alternative hardware. Using this ratio, \tool{} rescales the original power draw while considering power draw and speed of the alternate hardware.

\begin{figure*}[h]
  \centering
  \includegraphics[width=.7\linewidth]{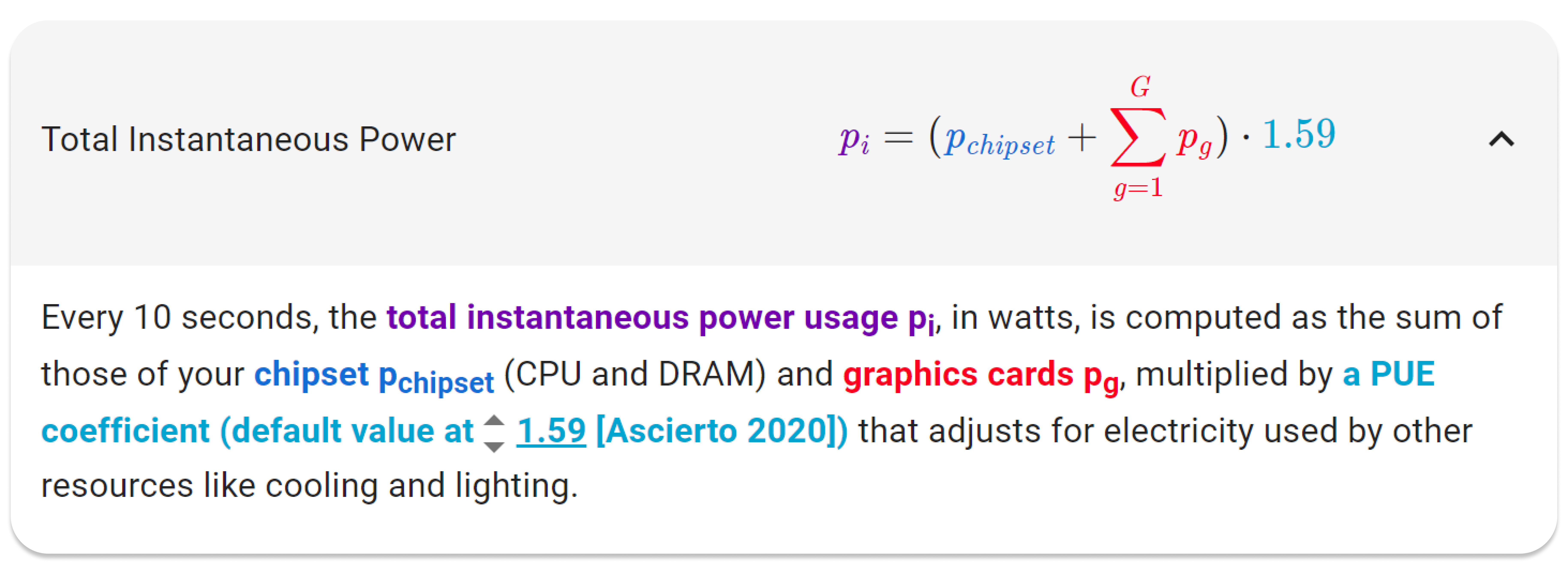}
  \caption{An example of an expanded color equation. The above equation computes instantaneous power usage of all hardware components. Users can adjust the PUE constant directly in the text above.}
  \Description{A component showing an expanded color equation. In this figure, a user has clicked on the color equation and can see text beneath the color equation. This text is also colored; similarly colored words correspond to similarly colored terms in the equation.}
  \label{fig:coloreq}
\end{figure*}

\subsubsection{Consumption Chart}
\label{consumption-chart}
The Consumption Chart component plots energy data from an energy profile alongside selected alternatives, for each epoch. Data from the selected energy profile is loaded into the chart component (left, Figure \ref{fig:graphmap}). The consumption chart also allows users to toggle between carbon emissions (CO$_{2}$) and electricity usage (kWh). \tool{} includes electricity usage since some research labs or organizations may be restricted when selecting region alternatives; a fairer comparison between models should rely on electricity usage instead of carbon emissions.

Users can also extrapolate beyond the provided data points in the selected energy profile. On the X-axis for the Consumption Chart (bottom left, Figure \ref{fig:graphmap}) users can use a stepper to increase or decrease extrapolated epochs. \tool{} performs a least-squares linear regression on the recorded data, and predicts carbon emissions or electricity usage for the extrapolated points. Given this data, users can make informed decisions about training models for longer periods of time.

\subsubsection{Color Equations} 
\label{color-eq-section}
To provide transparency into the calculations of carbon emissions, \tool{} provides dynamically changing equations that walk users through emission computations. Drawing from \citet{azad} and \citet{hohman2020communicating}, color equations highlight variables in color and provide corresponding text annotations in the same color. In \tool{}, color equations can be contracted (Figure \ref{fig:teaser}) or expanded (Figure \ref{fig:coloreq}); and allow users to directly edit constants within the explaining text, causing \tool{}'s remaining components to reflect the updated changes. Likewise, edits made in the Alternative Region component also affect constants in the color equations.

\tool{} utilizes 3 equations to compute carbon emissions (seen in \ref{fig:teaser}). The first equation computes CO$_{2}$ emissions from each epoch by multiplying electricity used over an epoch with a U.S state-dependent energy intensity constant (i.e. how much carbon each kilowatt hour of electricity produces). The second equation computes the electricity used during each epoch. Lastly, the third equation is described in Figure \ref{fig:coloreq}.

\subsection{\tool{} Live-Tracking Backend}

\label{backend-section}

\tool{}'s \textit{optional} live-tracking backend allows users to live-track and collect energy profiles for their models. If users already have energy profiles, \tool{} can be backend-agnostic. The backend from \tool{} is similar to \citet{anthony2020carbontracker}: it monitors CPU energy usage using Intel's Running Average Power Limit interface \cite{rapl_paper}, collects GPU usage through NVIDIA's System Management interface, and is written in Python. Like \citet{anthony2020carbontracker}, using the backend involves simply wrapping the model training code in a separate function provided by \tool{}. 

However, \tool{} contains modifications that allows the frontend to listen to training progress on the backend. On running the training code, \tool{} launches a backend server that exposes an API to the frontend, and prints the access URL to the console. The API provides a constantly updating model energy profile---users can enter the API's access URL into \tool{}'s frontend, enabling live-tracking.

\section{Usage Scenarios}

We envision the usage scenarios of EnergyVis to include user communities involved in intensive computational tasks, where information about environmental impacts may be difficult to access or visualize. Here, we highlight three usage scenarios focusing on ML.

\paragraph{\textbf{Tracking CO$_{2}$ emission in real-time, and comparing results:}}
Susan, a graduate student, is applying a series of models to a large image classification task. Susan visits a leaderboard website and selects the two promising models as candidates: EfficientNet \cite{efficientnet} and ResNet \cite{he2015deep}. Although Susan is unfamiliar with the specific techniques employed by the black-box models she is using, she \textit{is} aware of the potential energy efficiency limitations of her selected models; therefore, she wants to limit the electricity usage and emissions of her models. Susan cannot find an energy profile from her colleagues or online, so she begins actively tracking each of her models. she starts by using \tool{}'s \textbf{live-tracking} mode. Susan begins by wrapping her EfficientNet training loop with \tool{}'s Python plugin, and starts her training process---the training URL is printed to her console by \tool{}. She then loads \tool{}'s frontend, and enters the training URL in the Energy Profile component. She lets her model train for only 10 epochs, then halts the training process, since she cannot afford to spend too much time training. She then exports a model energy profile, and repeats this process for ResNet. After manually looking at the profiles, Susan realizes that the accuracy differences between EfficientNet and ResNet are insignificant for her task; however, EfficientNet uses significantly less energy. Susan confirms this suspicion by loading the profiles back into \tool{}, and extrapolating several epochs in the Consumption Chart. Susan finally decides to use EfficientNet for her task.

\paragraph{\textbf{Comparing electricity usage across different preloaded models:}}
James is a software engineer working on a sentiment analysis task. He's interested in utilizing a state-of-the-art model for his task, but he's unsure on what model he wants to use. Unfortunately, the startup he works for is trying to keep the monthly energy bill low, so he needs to be aware of energy efficiency. James opens \tool{} and notices that energy profiles have already been collected for two language models (a ``vanilla'' Transformer \cite{vaswani2017attention} model, and BERT \cite{devlin2019bert}) trained on a variant of his sentiment analysis task---because profiles have already been collected, James does not need to write any code, or use \tool{}'s backend! He selects the BERT model, then right clicks on the Transformer (Figure \ref{fig:energyprof}), loading the Transformer's data into the alternative views and overlaying the original carbon emissions with alternative emissions. James notices that \tool{} is currently displaying Carbon Emissions, so he toggles the Consumption Graph component to display electricity usage instead. He realizes that the Transformer uses less energy while having similar performance as BERT, and decides to use the Transformer for his task.

\paragraph{\textbf{Picking location and hardware alternatives: }}
Sarah, another graduate student, is working on a text classification task. She already designed a novel architecture, live-tracked her model, and collected energy profiles. However, she's interested in exploring alternatives to reduce her emissions when training her model, since she plans on tweaking and training repeatedly. She opens \tool{} and loads her model's energy profile. Next, she begins using the Model Region component to compare the $CO_2$ consumption in alternate states to her original state. Her research institution has compute deployed across several states. Sarah hovers over each lighter-colored state and compares the experiment consumption charts. After completing her search, she finds the least energy intensive state. Sarah also realizes that the state she selected has potentially more efficient hardware. To confirm this suspicion, she enters the hardware details into Alternative Hardware component, and notices that the alternative consumption in the Consumption Chart reduces again. Finally, she decides on deploying her model in her selected state.

\section{Ongoing Work \& Conclusion}

\paragraph{\textbf{Planned Evaluation.}}
We plan to extend our work by evaluating \tool{} using a two-phase user study. 
We will recruit researchers and practitioners as participants of the study, where they will use \tool{} for their ongoing research projects. 
Both phases may be conducted fully remotely via video-conferencing software (e.g., Zoom, BlueJeans). We are developing the user study protocol and will apply for our institution's IRB (institutional review board) approval. Given participant permission, we plan to record computer screens and microphone audio for later analysis. For both phases, we will provide participants with a list of \tool{}'s features, while encouraging participants to try them. Users will also be encouraged to think aloud and ask questions. All user study sessions will end with a questionnaire asking about \tool{}'s usability (e.g., Easy to use? Easy to understand?) and feature usefulness. %

Phase 1 will be a lab study whose focus is to evaluate \tool{}'s usability.
We will design a number of tasks that involve live-tracking a popular model's energy efficiency (like EfficientNet\cite{efficientnet} or a Transformer \cite{vaswani2017attention}), trained on a subset of ImageNet \cite{imagenet_cvpr09}, or a sentiment classification task. Models and datasets for our tasks will be chosen to reflect a realistic training context for participants. We also plan on creating tasks that involve identifying methods to reduce emissions through \tool{}'s alternative region or hardware components. Based on the user feedback from phase 1, we will improve \tool{}'s usability and features.

In phase 2,
we plan to evaluate the \textit{effectiveness} of \tool{} in raising awareness and potentially reducing emissions. 
Our participants will be asked to 
use \tool{} for their ongoing research projects for a month, while collecting feedback about their experience.
During this timespan, we will schedule regular, short check-ins (twice a week) with the participants to collect feedback about the tool's impact on their research, e.g., whether the tool leads to a change in their typical model training workflow, or prompts participants to reflect on their choices of hardware and cloud providers.
\paragraph{\textbf{Model Architecture Alternatives.}}
Currently, \tool{} offers alternatives for hardware and location based counterfactuals. However, architectural changes have also resulted in reduced energy consumption. Models like MobileNet \cite{howard2017mobilenets} and REST \cite{duggal2020rest} use techniques like compression and early exiting to reduce parameter counts and, therefore, energy consumption. Suggesting alternatives and estimating updated energy consumption based on these techniques might provide users with another avenue to reduce energy usage.

\paragraph{\textbf{Increased Region Support.}} Currently, \tool{} only supports region based alternatives for the United States. In future iterations, we aim to include locations that have readily available energy intensities (carbon to energy production values). The European Union's environment agency, for example, provides these intensity values---similar to NREL in the United States. 

\paragraph{\textbf{Deployment.}} \tool{} will eventually be open sourced for users to track their own models. We also plan on allowing users to easily share and import model energy profiles through pull requests on \tool{}'s codebase, so users can easily compare their own efficiency results with others in the research community.

\paragraph{\textbf{Conclusion.}} As models grow larger, identifying \textit{how} to reduce the environmental impact of these models will lead to fairer and more sustainable training pipelines. To this end, \tool{} provides an interactive means to explore the energy usage of ML models based on hardware and location of deployment. As we continue development on \tool{}, we aim to add increased region support, support model architecture alternatives, allow users to share energy profiles on \tool{} itself, and evaluate \tool{} through a user study.

\begin{acks}
This work was supported in part by NSF grants IIS-1563816, CNS-1704701, 1945332; DARPA (HR00112030001); gifts from Facebook, Intel, NVIDIA, Bosch, Google, Symantec, Yahoo! Labs, eBay, Amazon.
\end{acks}

\bibliographystyle{ACM-Reference-Format}
\bibliography{sample-base}

\end{document}